\documentclass[10pt,twocolumn,letterpaper]{article}
\usepackage{soul}
\usepackage{iccv}
\usepackage{times}
\usepackage{epsfig}
\usepackage{graphicx}
\usepackage{subfigure}
\usepackage{amsmath}
\usepackage{amssymb}
\usepackage{booktabs}
\usepackage{bbm}
\usepackage{bm}
\usepackage{multirow}
\usepackage{booktabs}
\usepackage[normalem]{ulem}

\usepackage[pagebackref=true,breaklinks=true,letterpaper=true,colorlinks,bookmarks=false]{hyperref}

\iccvfinalcopy 

\def\iccvPaperID{7555} 

\ificcvfinal\fi

\begin{document}

\title{CLIPN for Zero-Shot OOD Detection: Teaching CLIP to Say No}

\author{
Hualiang Wang, Yi Li, Huifeng Yao, Xiaomeng Li\thanks{Corresponding author.}\\
Department of Electronic and Computer Engineering, The Hong Kong University of Science and Technology\\
{\tt\small \{hwangfd, ylini, yhfpro, eexmli\}@ust.hk}\\
}

\maketitle
\ificcvfinal\thispagestyle{empty}\fi

\begin{abstract}
Out-of-distribution (OOD) detection refers to training the model on an in-distribution (ID) dataset to classify whether the input images come from unknown classes. Considerable effort has been invested in designing various OOD detection methods based on either convolutional neural networks or transformers. However, zero-shot OOD detection methods driven by CLIP, which only require class names for ID, have received less attention.
This paper presents a novel method, namely CLIP saying ``no'' (\textbf{CLIPN}), which empowers the logic of saying ``no" within CLIP. Our key motivation is to equip CLIP with the capability of distinguishing OOD and ID samples using positive-semantic prompts and negation-semantic prompts. Specifically, we design a novel learnable ``no" prompt and a ``no" text encoder to capture negation semantics within images. Subsequently, we introduce two loss functions: the image-text binary-opposite loss and the text semantic-opposite loss, which we use to teach CLIPN to associate images with ``no" prompts, thereby enabling it to identify unknown samples. Furthermore, we propose two threshold-free inference algorithms to perform OOD detection by utilizing negation semantics from ``no" prompts and the text encoder.
Experimental results on 9 benchmark datasets (3 ID datasets and 6 OOD datasets) for the OOD detection task demonstrate that CLIPN, based on ViT-B-16, outperforms 7 well-used algorithms by at least 2.34\% and 11.64\% in terms of AUROC and FPR95 for zero-shot OOD detection on ImageNet-1K. Our CLIPN can serve as a solid foundation for effectively leveraging CLIP in downstream OOD tasks. The code is available on \href{https://github.com/xmed-lab/CLIPN}{https://github.com/xmed-lab/CLIPN}.

\end{abstract}

\section{Introduction}
\label{sec:intro}

Deep learning models~\cite{he2016deep, dosovitskiy2020image} have demonstrated excellent versatility and performance when the classes in the training and test datasets remain the same~\cite{nguyen2015deep}. This is facilitated by the fact that the models are trained under completely closed-world conditions, meaning all encountered classes are in-distribution (ID) ones~\cite{drummond2006open}. Nevertheless, these models tend to suffer from poor generalization and undesirable performance when deployed in real-world applications. This frustrating phenomenon is partially attributed to the existence of an enormous number of \emph{unknown classes} distributed in the real world, which is challenging to detect as they have not been explicitly seen during the training stage. 

\begin{figure}[h]
  \centering
    \includegraphics[width=1.0\linewidth]{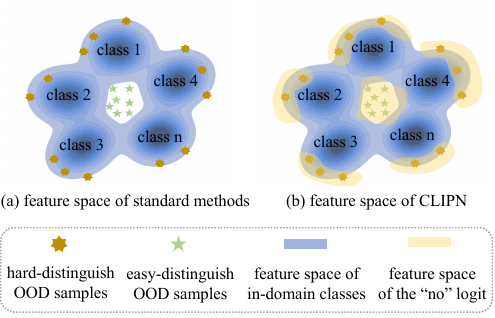}
   \caption{A toy comparison illustration of feature spaces between standard OOD detection algorithms and the proposed CLIPN. Our method involves a ``no" logic, which provides a new feature space (yellow region) to directly identify OOD samples. The qualitative experiment visualization is shown in Figure.~\ref{fig:tsne}. }
   \label{fig:1}
\end{figure}

Out-of-distribution (OOD) detection task~\cite{bendale2016towards, chen2017outlier, fort2021exploring} has been raised and promptly attracted considerable interest from researchers. Briefly, the OOD detection task aims to empower the model to distinguish if the input images come from unknown classes. 
One of the mainstream OOD detection methods is to learn ID-specific features and classifiers, then develop the scoring function~\cite{wang2022vim,hendrycks2016baseline} to metric how closely the input data matches the ID classes, which is measured by ID-ness~\cite{wang2022vim} (or OOD-ness, an opposite case). 
For instance, MSP~\cite{hendrycks2016baseline}, MaxLogit~\cite{hendrycks2019scaling}, energy-based~\cite{liu2020energy} and gradient-based~\cite{liang2017enhancing} have been extensively employed to measure the ID-ness.
Better ID-ness brings better OOD results.
In summary, the key idea of these methods is to \emph{teach ID knowledge to the model and then detect the dis-matched cases referring to the model's reply (score)}. 
The effectiveness of the above methods is seriously compromised by the following cases. 
As illustrated in Fig.~\ref{fig:1}, the green stars represent some OOD samples that are easy to distinguish, as they are  relatively distant from all ID classes and naturally have high entropy, uniform probability~\cite{hendrycks2016baseline}, low logit~\cite{hendrycks2019scaling} or low energy~\cite{liu2020energy}. 
Conversely, hard-to-distinguish OOD samples (brown stars in Fig.~\ref{fig:1}) are more common and challenging. 
These samples are located relatively close to a certain ID class while being far away from other classes, resulting in high ID-ness. Therefore, existing methods such as those mentioned above fail to identify such samples accurately.
As results shown in Fig.~\ref{fig:tsne}, even when we apply MSP~\cite{hendrycks2016baseline} with different thresholding, there are still numerous mis-classified OOD samples, which are located in close proximity to ID classes.

Recently, some methods have sought to address the issue of hard-to-distinguish OOD samples by leveraging generalizable representations learned by CLIP~\cite{gao2021clip}, an open-world language-vision model trained on datasets with enormous volumes, such as Laion-2B~\cite{schuhmann2022laion}. 
Naturally, this task extends to zero-shot OOD detection (ZS OOD detection)~\cite{esmaeilpour2022zero, mcm}, which employs language-vision models to detect OOD data without requiring training  on the ID dataset. 
ZOC~\cite{esmaeilpour2022zero} uses an additional text encoder to generate some candidate OOD classes not included in ID classes. 
Unfortunately, it is inflexible and unreliable when faced with a dataset containing a large number of ID classes, rendering it challenging to scale for large datasets such as ImageNet-1K~\cite{krizhevsky2017imagenet}. 
MCM~\cite{mcm} leverages the text encoder component of CLIP and ID class prompts to obtain a more representative and informative ID classifier, which in turn enhances the accuracy of ID-ness estimates. However, this method still neglects to address the challenge of dealing with hard-to-distinguish OOD samples and suffers from limited performance; see results in Table~\ref{tab:imagenet}.

\begin{figure}[h]
  \centering
    \includegraphics[width=1.0\linewidth]{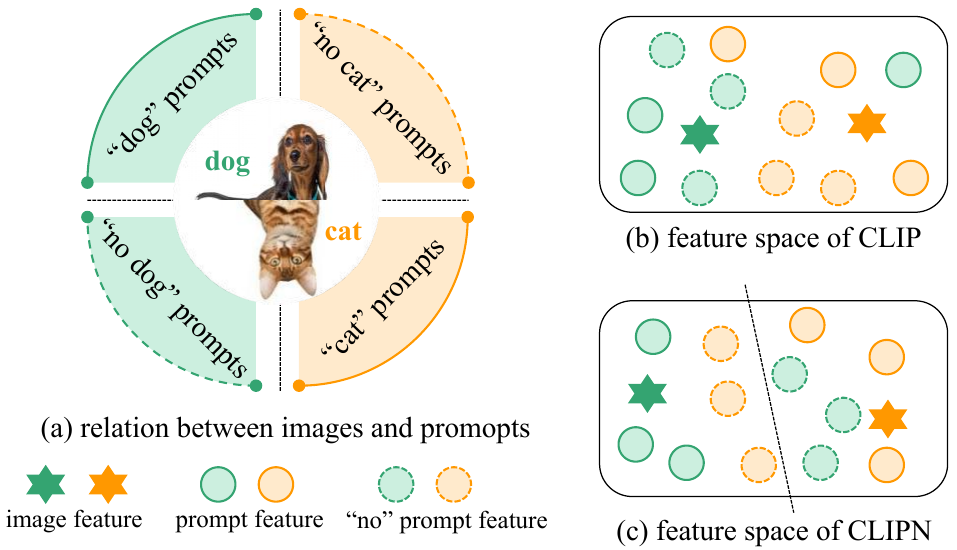}
   \caption{A toy illustration to determine that the original CLIP lacks ``no'' logic. The qualitative visualization is in Figure.~\ref{fig:dog_clipn}.
   } 
   \label{fig:2}
\end{figure}

Different from ZOC~\cite{esmaeilpour2022zero} and MCM~\cite{mcm}, we attempt to exploit the \textit{open-world} knowledge in CLIP to straightly \textit{identify} some hard-to-distinguish OOD samples even if their ID-ness is high.  
As the toy example shown in Figure.~\ref{fig:2} (a), given a dog image and a cat image, we design four groups of prompts. Two groups contain class prompts with/of/.../having the photos of the dog or cat, while the other two groups use ``no" prompts: a photo without/not of/.../not having the dog or cat. 
We conducted an experiment on CLIP to match the images with four prompts. Unfortunately, the results show that CLIP fails to  accurately match the images, implying that it lacks ``no'' logic; as illustrated in the toy visualization  in Fig.~\ref{fig:2} (b) and qualitative visualization in Fig.~\ref{fig:dog_clipn}. 

To empower ``no'' logic within CLIP, we propose a new CLIP architecture, called CLIP saying ``no'' (\textbf{CLIPN}). 
It upgrades CLIP in terms of OOD detection in three ways. 
\textbf{(1) Architecture}. New ``no'' prompts and a ``no'' text encoder are added to CLIP. Our novel learnable ``no'' prompts integrate negation semantics within prompts, complementing the original CLIP's prompts. 
Moreover, our ``no'' text encoder captures the corresponding negation semantics of images, making the CLIP saying ``no'' possible.  
\textbf{(2) Training Loss}. We further propose two loss functions. The first is image-text binary-opposite loss, which makes an image feature match with correct ``no'' prompt features. In other words, it can teach CLIP when to say ``no''.  The second is text semantic-opposite loss which makes the standard prompt and ``no'' prompts be embedded far away from each other. In other words, it can teach CLIP to understand the meaning of ``no'' prompts.
\textbf{(3) Threshold-free Inference Algorithms}. After the training of CLIPN, we design two threshold-free algorithms: competing-to-win and agreeing-to-differ. The goal of competing-to-win is to select the most confident probability from 
standard and ``no'' text encoders as the final prediction. While agreeing-to-differ generates an additional probability for the OOD class by considering predictions from both standard and ``no'' text encoders. 
Experimental results on 9 benchmark datasets (3 ID and 6 OOD datasets) showed that our CLIPN  outperforms existing methods.
In summary, our contributions are Four-fold. 
\begin{itemize}
    \item 
    We propose a novel CLIP architecture, named CLIPN, which equips CLIP with a ``no'' logic via the learnable ``no'' prompts and a ``no'' text encoder. 
   
    \item
    We propose the image-text binary-opposite loss and text semantic-opposite loss, which teach CLIPN to match images with ``no'' prompts, thus learning to identify unknown samples.
    
    \item We propose two novel threshold-free inference algorithms  (competing-to-win and agreeing-to-differ) to perform OOD detection via using negation semantics. 
    
    \item Experimental results show that our CLIPN outperforms most existing OOD detection algorithms on both large-scale and small-scale OOD detection tasks.  
    
\end{itemize}

\section{Related Work}

\label{sec:related}

\subsection{Contrastive Vision-Language Models}

One key research topic in artificial intelligence is studying the relationship between vision and language. Previously, many attention-based methods like BAN~\cite{kim2018bilinear}, Intra-Inter~\cite{gao2019dynamic}, and MCAN~\cite{yu2019deep} dominate the visual-language tasks. Then, inspired by BERT~\cite{devlin2018bert}, many mthods~\cite{lu2019vilbert, tan2019lxmert, petroni2019language} further contribute to the area via exploiting more advanced transformer architectures and prompt strategies. As transformer-based networks~\cite{dosovitskiy2020image, liu2021swin} are successfully deployed into computer vision tasks, CLIP~\cite{radford2021learning} is proposed to learn informative features from the vision-language pairs on top of the large-size model and datasets\cite{schuhmann2022laion} via contrastive learning~\cite{ding2021support}. Further, \cite{wortsman2022robust,xu2022groupvit} demonstrate its impressive generalization to downstream tasks~\cite{li2022exploring}. 
In addition, following the same lines as~\cite{petroni2019language}, there are also many works focusing on improving visual-language models via prompt engineering in computer vision.  
 CoOp~\cite{zhou2022learning} and CoCoOp~\cite{zhou2022conditional} work on changing the manual prompt to the learnable one for the purpose of better matching texts and images in a specific supervised task. 
Unlike these methods that provide more precise text descriptions or supervised guidance, this paper empowers the ``no'' logic within CLIP by adding learnable ``no'' prompts and text encoders during the unsupervised pre-trained period.
 


\subsection{CLIP-based Zero-Shot Learning}

Recently, zero-shot learning~\cite{huynh2020fine,wang2018zero} (ZSL) gains considerable interest. ZSL focuses on conferring the model with the ability to capture open (or unseen) knowledge. CLIP has contributed significantly to ZSL by virtue of CLIP's excellent open-set versatility~\cite{xu2022groupvit}. 
The representation space of CLIP is driven by a huge dataset~\cite{schuhmann2022laion} in an unsupervised manner and exhibits holistic, task-agnostic, and informative. These advantages~\cite{radford2021learning} of CLIP are the primary factors that have led to its outstanding zero-shot performance. Further, 
the zero-shot adaptability of CLIP is found to work across many domains and different practical tasks~\cite{liu2023chatgpt,wortsman2022robust,li2022freeseg,li2023clip}. 
ZOC~\cite{esmaeilpour2022zero} employs an additional image-text dataset and bert~\cite{devlin2018bert} module to learn the names of OOD classes. MCM~\cite{mcm} capture the powerful ID classifier from class prompts and get better ID-ness from CLIP models.
Unlike these methods, we exploit the open-world knowledge in CLIP  to design a new OOD detection pipeline, where uses CLIP to straightly identify OOD samples via negation-semantic prompts.

\subsection{Out-of-Distribution Detection}

The goal of OOD detection is to detect OOD images from the test dataset (containing both ID and OOD images). 
Designing the score function is the most popular method in OOD detection tasks. The scores are mainly derived from three sources: the probability, the logits, and the feature. For the probability, Hendrycks~\etal~\cite{hendrycks2016baseline} presented a baseline method using the maximum predicted softmax probability (MSP) as the ID score. Hendrycks~\etal~\cite{hendrycks2019scaling} also proposed to minimize KL-divergence between the softmax and the mean class-conditional distributions. For the logits, Hendrycks~\etal~\cite{hendrycks2019scaling} proposed using the maximum logit (MaxLogit) method. The energy score method~\cite{liu2020energy} proposed to compute the logsumexp function over logits. For the feature, Lee~\etal~\cite{lee2018simple} computed the minimum Mahalanobis distance between the feature and the class centroids. Ndiour~\etal~\cite{ndiour2020out} used the norm of the residual between the feature and the pre-image of its low-dimensional embedding. 
Furthermore, \cite{bai2023effectiveness} proposes a solution that utilizes contrastive loss to distinguish between ID and OOD samples. In contrast to the methods mentioned earlier, which solely rely on the concept of ID-ness derived from images, our CLIPN can directly identify OOD samples by considering both images and negation-semantic texts.

\section{Methodology}
\label{sec:method}

\subsection{Preliminary: CLIP-based OOD Detection} 
\label{sec:method_preliminary}

Profiting from massive volumes of training data and large-size models, CLIP has demonstrated impressive performance in the zero-shot out-of-distribution classification task. Naturally, exploring the potential of CLIP on zero-shot out-of-distribution detection is worthwhile. 

Referring to the existing CLIP-based zero-shot tasks~\cite{gao2021clip, mcm}, we briefly review how to perform zero-shot OOD detection.
The image encoder of CLIP is used to extract the image feature. Although there is no classifier in CLIP, we can use the names of ID classes to build the text inputs (e.g., ``a photo of the dog''). Then obtain the text features from the text encoder as the class-wise weights that functionally play the same role as the classifier.
Finally, taking the maximum softmax probability (MSP) algorithm as an example, we can calculate MSP according to the image feature and class-wise weights. The image will be treated as OOD if MSP is smaller than a pre-defined threshold, and vice versa.

\subsection{Overview of CLIPN}
\label{sec:method_2}
\begin{figure*}[h!]
  \centering
    \includegraphics[width=0.8\linewidth]{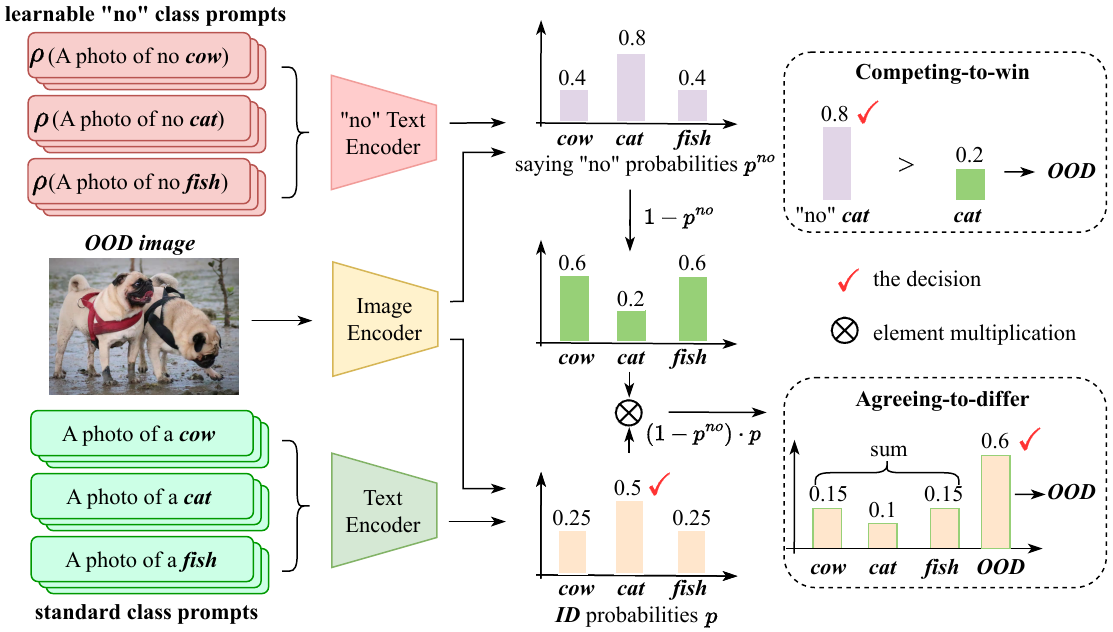}
   \caption{The inference pipeline of CLIPN. It consists of three networks: the image encoder, text encoder and ``no'' text encoder with \textbf{learnable ``no'' prompts} $\rho$. For the inference period, we propose competing-to-win and agreeing-to-differ to jointly determine the result with two text encoders. The ID classes are cow, cat, fish, and the OOD class is dog here.}
   \label{fig:clip_clipyn}
\end{figure*}

In this paper, we customize a new CLIP architecture, named CLIP saying ``no" (CLIPN), to exploit the potential of OOD detection in the original CLIP.

\noindent \textbf{Architecture.} As shown in Figure~\ref{fig:clip_clipyn}, our CLIPN is built on top of a pre-trained CLIP and consists of three modules:
\textbf{(1)Image Encoder: $\phi$.} 
The image encoder $\phi$ of CLIPN keeps the same structure and parameters as the image encoder of the pre-trained CLIP. In this paper, ViT-B~\cite{dosovitskiy2020image} and ViT-L~\cite{dosovitskiy2020image} are deployed to instantiate $\phi$, respectively. By default, the parameters of $\phi$ are frozen.
 \textbf{(2) Text Encoders: $\psi$.} 
The text encoder $\psi$ of CLIPN keeps the same structure and parameters as the text encoder of the pre-trained CLIP. The input of $\psi$ also keeps the same, i.e., a standard text that describes one image (e.g., ``a photo with/of/...'').
\textbf{(3)``no'' Text Encoders: $\psi^{no}$.} It is initialized by the text encoder of the pre-trained CLIP. But the difference from the $\psi$ is that we set $\psi^{no}$ learnable. The input of $\psi^{no}$ is a negative text that describes one image with the opposite semantic.

Besides, we access the pre-trained CLIP from the OpenCLIP~\cite{ilharco_gabriel_2021_5143773}. In this paper, we use two pre-trained models, including the CLIP based on ViT-B and ViT-L pre-trained using Laion-2B dataset~\cite{schuhmann2022laion}, respectively.

\noindent \textbf{Pre-training CLIPN.} Obviously, the input texts of $\psi$ and $\psi^{no}$ are the opposite semantic in terms of one input image, just like the opposite attribute of ID and OOD. To teach CLIPN to distinguish which semantic is correctly matched with the image, we first design a novel ``no'' prompt strategy which introduces the textual descriptions with negative logic into $\psi^{no}$; see Sec.~\ref{subsec:promptdesign} for details. 

Based on the above-mentioned thought of opposite semantic, we propose an image-text binary-opposite loss and text semantic-opposite loss (See Sec.~\ref{subsec:bocl} for details) which allow CLIPN to learn explicitly when and how to align images with two kinds of texts.

\noindent \textbf{Inference Stage.} In the inference period, we further propose two novel threshold-free algorithms to determine if the input image is OOD (See Sec.~\ref{subsec:infer} for details).
The first is named the competing-to-win algorithm. It is summarized as follows: we first use the standard text and the input image to predict the ID probabilities in terms of ID classes. Then employ the ``no'' text of the class with the highest probability to determine if the ID prediction is correct.
The second is named the agreeing-to-differ algorithm. We use ``no'' texts to shrink the ID probabilities and generate a new probability of the OOD class. If the OOD probability is highest, the input image will be judged as OOD.

\subsection{Prompt Design}
\label{subsec:promptdesign}

We propose a new prompting strategy in the unsupervised pre-trained period: learnable ``no'' prompt pools. 
In the original CLIP, the image is depicted in positive logic, i.e., the content of the text is semantically consistent with the content of the image. We attempt to supplement a kind of negative logic, i.e., the content of the text is semantically negative with the content of the image. To this end, we define a series of ``no'' prompts to equip the original texts. 
Denoting the text for one image $\bm{x}$ is $\bm{t}$, the ``no'' prompt pool is defined as ${pool}_{no}(\bm{t})$ = $\{$``a photo without \{$\bm{t}$\}'', ``a photo not appearing \{$\bm{t}$\}'', ..., ``a photo not containing \{$\bm{t}$\}''$\}$, where has $L$ handcrafted ``no'' prompts. The handcrafted ``no'' prompt pool is built upon~\cite{gao2021clip} and modified with negative keywords. During training, given the input text $\bm{t}$, we randomly pick a ``no'' prompt from ${pool}_{no}$ to generate the text with ``no'' logic, $\bm{t}^{no}$. Next, a token embedding layer $\rho$ is used to embed the ``no'' texts as a group of token feature vectors $\rho(\bm{t}^{no})$, which are the input tokens of the ``no'' text encoder. 

Moreover, inspired by~\cite{zhou2022conditional}, we design the learnable ``no'' prompts. We replace the negative keywords (e.g., ``a photo without'') by some learnable parameters $\sigma$ that present the token features of negative semantics and are combined with the token features of $\bm{t}$ in the feature space.

\subsection{Training Loss Design}
\label{subsec:bocl}
\begin{figure}[h!]
  \centering
    \includegraphics[width=0.9\linewidth]{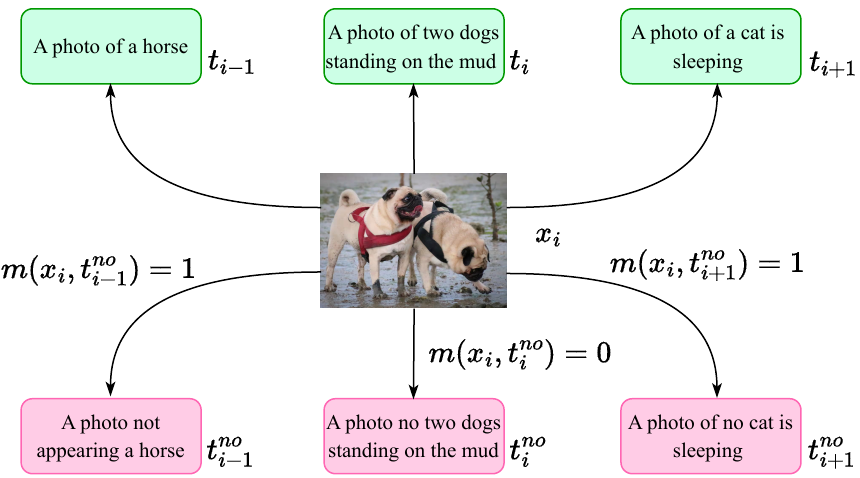}
   \caption{The illustration for matching $\bm{x}$ and $\bm{t}^{no}$. The green and pink boxes indicate the standard text $\bm{t}$ and ``no'' text $\bm{t}^{no}$, respectively. $m(\bm{x}_i, \bm{t}^{no}_j) = 1$ indicates they are \textit{matched yet unrelated} (i.e., the ``no'' text is not a wrong description yet semantically irrelevant). $m(\bm{x}_i, \bm{t}^{no}_j) = 0$ indicates they are \textit{reversed matched} (i.e., the ``no'' text has opposite semantic to the image ). } 
   \label{fig:demo_yij}
\end{figure}

In the training period, we define a mini-batch with $N$ input pairs as $\mathcal{B} = \{ (\bm{x}_i, \bm{t}_i, \bm{t}_{i}^{no})  \}_{i=1}^{N} \in \mathbb{D}_{clip}$ where $(\bm{x}_i, \bm{t}_i, \bm{t}_{i}^{no})$ is the $i$-th input pair including an image, a standard text, and a ``no'' text. $\mathbb{D}_{clip}$ is the dataset to pre-train CLIPN, such as CC12M~\cite{sharma2018conceptual}. Then the image feature $\bm{f}_i$, text feature $\bm{g}_i$ and ``no'' text feature $\bm{g}_{i}^{no}$ can be calculated as follows:
\begin{equation}
	\begin{split}
	\bm{f}_i &= \phi(\bm{x}_i)\\
	\bm{g}_i &= \psi(\bm{t}_i)\\
    \bm{g}^{no}_i &= \psi^{no}(\bm{t}^{no}_i),
	\end{split}
\label{eqn:1}
\end{equation}
where $\bm{f}_i, \bm{g}_i, \bm{g}^{no}_i  \in \mathbb{R}^{1 \times D}$ and $D$ is the feature dimension.  All features are normalized by $L_2$ normalization operation. 

We train CLIPN through two loss functions. 

\noindent \textbf{Image-Text Binary-Opposite Loss (ITBO).}
This loss function assists the model in matching the image feature with the correct ``no'' text features. To be specific, we define two relations between images and ``no'' texts: (1) matched yet unrelated (i.e., the ``no'' text is not the wrong description yet semantically irrelevant to the image); (2) reversed matched (i.e., the ``no'' text has opposite semantic to the image), as explained in Fig.~\ref{fig:demo_yij}. Consequently, the match-ness $m(\bm{x}_i, \bm{t}^{no}_j)$ between the $i$-th image and the $j$-th ``no'' text can be defined as follow:
\begin{equation}
	m(\bm{x}_i, \bm{t}^{no}_j) = m_{ij} = \begin{cases}
	0, & i = j, \\
	1, & i \neq j,
		   \end{cases}
\label{eqn:m}
\end{equation}
where $m(\bm{x}_i, \bm{t}^{no}_j) = 0$ indicates they are reversed matched and $m(\bm{x}_i, \bm{t}^{no}_j) = 1$ indicates they are matched yet unrelated.
Then we drive CLIPN to match images and ``no'' texts in the feature space, guided by the match-ness. The loss $\mathcal{L}_{itbo}$ is formulated as:
\begin{equation}
\begin{split}
	\mathcal{L}_{itbo}(\mathcal{B}) &= - \frac{1}{N} \sum_{i=1}^{N}  (1-m_{ii}) \log (1-p_{ii}^{no} )      \\
                    & - \frac{1}{N(N-1)} \sum_{i=1}^{N} \sum_{ j\neq i }^{N} m_{ij} \log p_{ij}^{no},
\end{split}
\end{equation}
where $p_{ij}^{no}$ presents the matched probability between the $i$-th image and $j$-th ``no'' text, formulated as follow:
\begin{equation}
p_{ij}^{no} = \frac{ e^{ <\bm{f}_i, \bm{g}^{no}_{j}> / \tau}  }{ e^{ <\bm{f}_i, \bm{g}_{j}> / \tau} + e^{ <\bm{f}_i,  \bm{g}^{no}_{j}> / \tau} },
\label{pno}
\end{equation}
where $<,>$ indicates the inner product of two vectors and $\tau$ is a learnable temperature parameter. 

\noindent \textbf{Text Semantic-Opposite Loss (TSO).} Obviously, the ``no'' prompts render $\bm{t}_i$ and $\bm{t}_{i}^{no}$ semantically opposite. Hence, in the feature space, the $\bm{g}_i$ and $\bm{g}_{i}^{no}$ also should be far from each other. To this end, the text semantic-opposite loss $\mathcal{L}_{tso}$ is defined as:
\begin{equation}
    \mathcal{L}_{tso}(\mathcal{B}) = \frac{1}{N} (2 -  \sum_{i=1}^{N} \Vert \bm{g}_{i} - \bm{g}_{i}^{no}   \Vert_2), 
\end{equation}
where $\Vert \Vert_2$ is the $L_2$ distance function. When all $\bm{g}_i$ and $\bm{g}_{i}^{no}$ pairs are embedded into the opposite directions in the feature space, $\mathcal{L}_{tso}(\mathcal{B})$ will decrease to $0$.
The total loss in a mini-batch is calculated by adding the above two loss values: $\mathcal{L}(\mathcal{B}) = \mathcal{L}_{itbo}(\mathcal{B}) + \mathcal{L}_{tso}(\mathcal{B})$.

\subsection{Inference algorithm of CLIPN}
\label{subsec:infer}
When deploying CLIPN to the test datasets of $\mathbb{D}_{id}$ and $\mathbb{D}_{ood}$ to perform a zero-shot OOD detection task, we only need all class names.
For the $C$ ID classes task,
as shown in Figure~\ref{fig:clip_clipyn}, the name of each class is feed into standard and ``no'' texts, then fed into $\psi, \psi^{no}$. Actually, we introduce the prompt pools in Sec.~\ref{subsec:promptdesign}. A prompt ensemble strategy is employed to generate the text features via adding the text features of all prompts together, followed~\cite{gao2021clip}
Then the ID class probability can be formulated as:
\begin{equation}
        p_{ij} = \frac{  e^{ <\bm{f}_i, \bm{g}_{j}> / \tau  }    }{  \sum_{k=1}^{C} e^{ <\bm{f}_i, \bm{g}_{k}>  / \tau  }       },
\end{equation}
where $C$ is the number of classes and $p_{ij}$ presents the predicted probability for the input image $x_i$ belonging to the $j$-th class. The matched probability $p_{ij}^{no}$ between the image and the $j$-th ``no'' class text can be calculated via Eqn.~\ref{pno}. 
Next, we propose two novel threshold-free algorithms to determine if $x_i$ is OOD. 

\noindent \textbf{Competing-to-win Algorithm.} The first algorithm is named the competing-to-win (CTW). Motivated by MSP, we find the class with the maximum ID probability. Then we compare the value of $p^{no}$ and $p^{yes}$ (i.e., $1-p^{no}$) to determine if $x_i$ is OOD (or ID). The above process can be formulated as follows:
\begin{equation}
	\begin{split}
	 {I}_{ctw} &= 1 - p_{ij}^{no}, j= \arg \max \{ p_{ik} \}_{k = 1 \sim C } \\
    \mathbb{I}(\bm{x}_i) &= \begin{cases}
	1, & 1-p_{ij}^{no} \geq p_{ij}^{no}, \\
	0, & else,
		   \end{cases}
	\end{split}
	\label{eq:idness}
\end{equation}
where $j$ indicates the class with the highest ID probability, ${I}_{ctw}$ is the ID-ness, and $\mathbb{I}(\bm{x}_i)$ is an indicator that presents $\bm{x}_i$ is an ID image when it is equal to 1, while is OOD .

\noindent \textbf{Agreeing-to-differ Algorithm.} Nevertheless, the above strategy is slightly aggressive. It will fail in some hard-distinguish situations (e.g., the maximum ID probability is not significantly higher than other probabilities). To make the decision flexible, 
we propose another algorithm, named agreeing-to-differ (ATD), to take all ID probabilities and $p^{no}$ into account. 
It can re-formulate the $C$-classes probabilities as the $(C+1)$-classes probabilities. An unknown class will be created, and its probability is defined as:
\begin{equation}
        p_{C+1} = 1 - \sum_{j=1}^{C} (1-p^{no}_{ij}) p_{ij}.
\end{equation}

The OOD sample can be detected as:
\begin{equation}
	\begin{split}
	I_{atd} &= 1 - p_{C+1} \\
  \mathbb{I}(\bm{x}_i) &= \begin{cases}
	1, & p_{C+1} \leq \max\{p_{ij}\}_{j=1 \sim C}. \\
	0, & else.
		   \end{cases}
	\end{split}
	\label{eq:idness_2}
\end{equation}

If the probability of the unknown class $p_{C+1}$ is larger than all ID probabilities, the input image will be detected as OOD. Otherwise, it is determined as ID.

\section{Experiment}
\label{sec:exp}

\subsection{Experimental Details}

\noindent \textbf{ID and OOD Datasets.} In this section, we evaluate the performance of our approach and compare it to state-of-the-art OOD detection algorithms. We focus on three different OOD detection tasks. 

(1) OOD detection on large-scale datasets. Following the prior work~\cite{mcm} on large-scale OOD detection, we choose ImageNet-1K~\cite{krizhevsky2017imagenet} as the ID dataset. Four OOD datasets (including Texture~\cite{wang2022vim}, iNaturalist~\cite{van2018inaturalist}, SUN~\cite{xiao2010sun}, and Places365~\cite{zhou2017places}) are used to comprehensively benchmark the algorithms. 

(2) OOD detection on small-scale datasets. In this setting~\cite{koner2021oodformer}, CIFAR-100~\cite{krizhevsky2009learning} is set as an ID dataset, and the OOD datasets are CIFAR-10~\cite{krizhevsky2009learning}, ImageNet\_R~\cite{koner2021oodformer}, and LSUN~\cite{koner2021oodformer}. The data scale of CIFAR-100 is significantly smaller than ImageNet-1K.

(3) OOD detection on the in-domain dataset. In addition, ZOC~\cite{esmaeilpour2022zero} is the first work on the zero-shot OOD detection task. To compare with it, we follow its experimental setting and test our method on CIFAR-10 and CIFAR-100 datasets. The reported performance is averaged over 5 splits.

\begin{table*}[h]
    \centering
    \caption{Results on large-scale zero-shot OOD detection. ID dataset is ImageNet-1K. All experimental results are evaluated using CLIP model based on ViT-B-16 and ViT-B-32. CLIPN-C and CLIPN-A indicate CLIPN with competing-to-win and agreeing-to-differ algorithms, respectively. The number with \textcolor{red}{red} and \textcolor{green}{green} indicates the improved AUROC and FPR95 of our algorithms compared to the baseline algorithms, respectively. The \textbf{bold} number presents the best performance on each dataset in terms of AUROC and FPR95. The result of MCM is reported on \cite{mcm} and the results of other compared methods are reproduced by us.}
    \label{tab:imagenet}
\resizebox{1.0\linewidth}{!}{
\begin{tabular}{c|cccccccc|cc}
\toprule
\multirow{2}{*}{Method} & \multicolumn{2}{c}{iNaturalist} & \multicolumn{2}{c}{SUN} & \multicolumn{2}{c}{Texture} & \multicolumn{2}{c|}{Places} & \multicolumn{2}{c}{\textbf{Avg}} \\
\cline{2-11}
                  & AUROC$\uparrow$ & FPR95$\downarrow$    & AUROC$\uparrow$ & FPR95$\downarrow$   & AUROC$\uparrow$ & FPR95$\downarrow$      & AUROC$\uparrow$  &FPR95$\downarrow$ & AUROC$\uparrow$  &FPR95$\downarrow$\\
\midrule
\multicolumn{11}{c}{Image Encoder: ViT-B-16} \\
\midrule
MSP~\cite{hendrycks2016baseline}               & 77.74           & 74.57       & 73.97          & 76.95             & 74.84        & 73.66         & 72.18          & 79.72           & 74.68       &   76.22       \\
MaxLogit~\cite{hendrycks2019scaling}          & 88.03     & 60.88            & 91.16   & 44.83            & 88.63     &  48.72    & 87.45    & 55.54      & 88.82        & 52.49      \\
Energy~\cite{liu2020energy}            & 87.18           & 64.98            & 91.17         & 46.42            &  88.22      & 50.39         & 87.33           & 57.40            & 88.48          & 54.80          \\
ReAct\cite{sun2021react}      & 86.87             & 65.57            & 91.04           & 46.17       & 88.13  &   49.88  & 87.42           & 56.85             & 88.37    & 54.62          \\
ODIN~\cite{liang2017enhancing}      & 57.73   & 98.93  & 78.42   & 88.72   &    71.49    &    85.47       & 76.88           & 87.8       & 71.13          & 90.23            \\
MCM~\cite{mcm} & 94.61     & 30.91           & 92.57     & 37.59          & 86.11           & 57.77            & 89.77     & 44.69         & 90.76           &  42.74         \\
\midrule
CLIPN-C (Ours)             & 90.88    & 28.58      & 89.38  & 31.64        & 78.28    & 56.59       &  86.85  & 37.55    & 86.35  & 38.59       \\
CLIPN-A (Ours)         & \textbf{95.27}  & \textbf{23.94}    & \textbf{93.93}   & \textbf{26.17}   & \textbf{90.93}      & \textbf{40.83}      & \textbf{92.28}    & \textbf{33.45}      & \textbf{93.10} (\textcolor{red}{+2.34})  & \textbf{31.10} (\textcolor{green}{-11.64})   \\ 
\midrule
\multicolumn{11}{c}{Image Encoder: ViT-B-32} \\
\midrule
MSP~\cite{hendrycks2016baseline}               & 77.14           & 74.62       & 71.86           & 80.64             & 73.76         & 74.88         & 70.61           & 82.09             & 73.34         &   78.06         \\
MaxLogit~\cite{hendrycks2019scaling}          & 84.01     & 75.05             & 88.27    & 56.59             & 84.41     &  60.00    & 85.68     & 61.65       & 85.59           & 63.32       \\
Energy~\cite{liu2020energy}            & 82.51           & 80.36             & 88.14           & 59.96             &  83.66      & 63.18           & 85.44           & 64.61             & 84.94           & 67.03            \\
ReAct\cite{sun2021react}      & 84.00             & 76.94             & 88.05           & 59.27       & \textbf{88.16}  &   53.31  & 85.41           & 64.42             & 86.41     & 63.49            \\
ODIN~\cite{liang2017enhancing}      & 44.57  & 99.60  & 76.63   & 94.39  &    71.85  &    89.88      & 75.84          & 91.90       & 67.22         & 93.94         \\
\midrule
CLIPN-C (Ours)             & 87.24     & 38.42       & 84.49  & 43.43    & 72.93    & 65.41     &  76.26  & 60.73	    & 80.23	 & 52.00       \\
CLIPN-A (Ours)         & \textbf{94.67}  & \textbf{28.75}   & \textbf{92.85}     & \textbf{31.87}    & 86.93 & \textbf{50.17}       & \textbf{87.68}    & \textbf{49.49}     & \textbf{90.53} (\textcolor{red}{+4.12})  & \textbf{40.07} (\textcolor{green}{-23.25})   \\ 
\bottomrule
\end{tabular}}
\end{table*}

\begin{table*}[h]
    \centering
    \caption{Results on small-scale zero-shot OOD detection. ID dataset is CIFAR-100. The image encoder is ViT-B-32.}
    \label{tab:cifar100_zero}

    \resizebox{1.0\linewidth}{!}{
\begin{tabular}{c|cccccc|cc}
\toprule
\multirow{2}{*}{} & \multicolumn{2}{c}{CIFAR-10} & \multicolumn{2}{c}{ImageNet\_R} & \multicolumn{2}{c|}{LSUN} & \multicolumn{2}{c}{Avg} \\
\cline{2-9}
                  & AUROC$\uparrow$        & FPR95$\downarrow$        & AUROC$\uparrow$         & FPR95$\downarrow$         & AUROC$\uparrow$          & FPR95$\downarrow$             & AUROC$\uparrow$        & FPR95$\downarrow$          \\
\midrule
MSP~\cite{hendrycks2016baseline}               & 77.83           & 80.02             & 85.09           & 58.42       & 82.51           & 68.88             & 81.81           & 69.11             \\
MaxLogit~\cite{hendrycks2019scaling}          & 85.20     & 57.58       & 78.79           & 83.72             & 89.02           & 61.21             & 84.34           & 67.51             \\
Energy~\cite{liu2020energy}            & 84.04           & 59.16             & 74.65           & 88.81             & 87.37           & 68.87             & 82.02           & 72.28             \\
ReAct\cite{sun2021react}      & 83.32           & 61.13             & 75.24           & 87.88             & 88.59           & 63.12             & 82.38           & 70.71             \\
ODIN~\cite{liang2017enhancing}       & 69.38         & 91.57          & \textbf{87.30}          & \textbf{48.65}          &   93.32        &     \textbf{32.21}     & 83.33         & 57.48            \\
\midrule
CLIPN-C (Ours)               & 85.44    & 49.84	  &  82.72	  & 65.79        & 	91.51    & 44.26        & 86.56      & 53.30      \\
CLIPN-A (Ours)            & \textbf{88.06}  & \textbf{47.99}       & 87.09    & 60.07 & \textbf{93.55}     & 35.19   & \textbf{89.57} (\textcolor{red}{+5.23})  & \textbf{47.75} (\textcolor{green}{-9.73})   \\ 
\bottomrule
\end{tabular}}
\end{table*}

\noindent \textbf{Evaluation Metrics.}
We report two widely used metrics, AUROC and FPR95, for performance evaluation. AUROC is a metric that computes the area under the receiver operating characteristic curve. A higher value indicates better detection performance. FPR95 is short for FPR@TPR95, which is the false positive rate when the true positive rate is 95\%. Smaller FPR95 implies better performance.

\noindent \textbf{Model Details.}
All used pre-trained CLIP models are obtained from OpenCLIP~\cite{ilharco_gabriel_2021_5143773}. The CLIP based on ViT-B-16 and ViT-B-32 are used for experiments.

\noindent \textbf{Previous Methods for Comparison.}
We compare our method with 5 prior logit-based works on OOD detection and 2 works on zero-shot OOD detection.
They are MSP~\cite{hendrycks2016baseline}, Energy~\cite{liu2020energy}, MaxLogit~\cite{hendrycks2019scaling},  ReAct~\cite{sun2021react}, ODIN~\cite{liang2017enhancing}, ZOC~\cite{esmaeilpour2022zero} (only on the third task) and MCM~\cite{mcm}. 
For evaluation, ID-ness socres proposed by the above works and this paper (see Eqn.~\ref{eq:idness} and \ref{eq:idness_2}) are used for the calculation of AUROC and FPR95. 

\noindent \textbf{Reproduction details.}
CLIPN is pre-trained on the CC-3M~\cite{sharma2018conceptual} dataset for ten epochs, using a batch size of 2048, and trained on a system equipped with $4\times$NVIDIA RTX3090 (24G) and $2\times$Intel Xeon Gold 5118.
All experiments involving our proposed algorithms are independently repeated three times, and the reported results are averaged from these three experiments. To ensure fairness and versatility in our evaluation, the model used for performance assessment is the final-epoch model, rather than selecting the optimal model from all epochs.

\subsection{Results on Zero-shot OOD Detection }
\noindent \textbf{OOD detection on large-scale datasets.}
Large-scale OOD detection contributes to real-world applications. We give a comprehensive OOD evaluation on ImageNet-1K in Table~\ref{tab:imagenet}. 
Our proposed method, CLIPN-A with learnable 'no' texts, outperforms the previous state-of-the-art method MCM~\cite{mcm} across all OOD datasets, achieving the highest AUROC and FPR95 scores. On average, our approach using ViT-B-16 demonstrates significant enhancements of at least 2.34\% and 11.64\% in terms of AUROC and FPR95, respectively. Similarly, our approach utilizing ViT-B-32 showcases improvements of at least 4.12\% and 23.25\% on AUROC and FPR95, respectively, underscoring its superior performance.

%

\noindent \textbf{OOD detection on small-scale datasets.}  In Table~\ref{tab:cifar100_zero}, we present the results of our method on CIFAR-100. Our approach achieves the best average performance on small datasets in terms of AUROC and FPR95.  Overall, our experiment yielded an impressive conclusion: our algorithms demonstrate superior generalizability, consistently achieving optimal average performance for zero-shot OOD detection tasks of any scale.


\begin{table}[t]
\caption{Results of OOD detection on the in-domain setting. $\dag$ and $\ddag$ indicate that the results are evaluated based on pre-trained model in ZOC~\cite{esmaeilpour2022zero} and pre-trained model in our paper. $\vartriangle$ indicates the improved AUC and FPR95.}
\label{tab:in_domain}

\resizebox{1\columnwidth}{!}{
\begin{tabular}{c|cccc}
\toprule
          & \multicolumn{2}{c}{CIFAR10} & \multicolumn{2}{c}{CIFAR100}  \\
          & AUROC           & FPR95     & AUROC             & FPR95     \\
\midrule
CLIP+MSP~$\dag$~\cite{hendrycks2016baseline}  & 88.0$\pm$3.3   & -          & 78.1$\pm$3.1      & -         \\
ZOC~$\dag$~\cite{esmaeilpour2022zero}       & \textbf{93$\pm$1.7}  & -         & 82.1$\pm$2.1              & -        \\
$\vartriangle$  & \textcolor{red}{+5} & - & \textcolor{red}{+4} & -    \\
\midrule
CLIP+MSP~$\ddag$~\cite{hendrycks2016baseline} & 85.58$\pm$2.83       & 68.68$\pm$8.86     & 74.76$\pm$4.42          & 84.29$\pm$6.78    \\
CLIPN-C~$\ddag$   & 89.15$\pm$1.16       & 53.43$\pm$2.07     & 81.29$\pm$2.51          & 66.91$\pm$5.58    \\
CLIPN-A~$\ddag$  & 90.08$\pm$1.14       & 53.21$\pm$2.53     & \textbf{82.51$\pm$1.78} & 65.98$\pm$5.34    \\
$\vartriangle$  & \textcolor{red}{+4.5} & \textcolor{green}{-15.47} & \textcolor{red}{+7.75} & \textcolor{green}{-18.31}    \\
\bottomrule
\end{tabular}}
\end{table}

\noindent \textbf{OOD detection on in-domain datasets.} Table.~\ref{tab:in_domain} shows our results in the in-domain setting, where CIFAR-10 and CIFAR-100 are divided into ID groups and OOD groups, respectively. 
It's worth noting that our method outperforms MSP by 4.5\% and 7.75\% AUROC on CIFAR10 and CIFAR100, respectively. Meanwhile, ZOC also outperforms MSP by 5\% and 4\% AUROC. However, it should be pointed out that our implemented MSP yielded lower scores than those reported in ZOC~\cite{esmaeilpour2022zero}, which is caused by the higher-performance pre-trained model (but no public available pre-tarined model of ZOC).


\subsection{Ablation Study}

\noindent \textbf{The effectiveness of two losses.}
We conduct experiments to demonstrate the effectiveness of the proposed two loss functions, image-text binary-opposite (ITBO) loss and text semantic-opposite (TSO) loss. 
We train CLIPN using ITBO and ITBO+TSO, respectively. 
The results evaluated by CLIPN-A with learnable ``no'' texts are reported in Table.~\ref{tab:abloss}.
When solely utilizing ITBO, CLIPN-A achieves AUROC scores of 93.08\%, 91.61\%, 89.52\%, and 89.83\% across the four OOD datasets.
Upon incorporating both ITBO and TSO, an additional enhancement of 2.19\%, 2.32\%, 1.41\%, and 2.45\% is observed in the AUROC scores.

\begin{table}[t]
    \centering
    \caption{Ablation study on the proposed loss. The reported AUC is evaluated by CLIPN-A with learnable ``no'' texts. ITBO and TSO indicate image-text binary-opposite loss and text semantic-opposite loss.}
    \label{tab:abloss}

\begin{tabular}{c|cccc}
\toprule
 Loss & iNaturalist & SUN & Texture & Places  \\
\midrule
ITBO  &  93.08    &    91.61    &  89.52    &      89.83      \\
ITBO + TSO  &  95.27    &    93.93   &    90.93     &      92.28 \\
\bottomrule
\end{tabular}
\end{table}

\noindent \textbf{The effectiveness of the handcrafted and learnable ``no'' texts.}
We conducted experiments to demonstrate the effectiveness of both handcrafted and learnable ``no'' texts. The handcrafted ``no'' texts are constructed by modifying the text prompts of CLIP using negative keywords. The number of learnable ``no'' keywords is set to 16. The results evaluated using CLIPN-C and CLIPN-A are presented in Table.~\ref{tab:prompt}. For CLIPN-C, the utilization of learnable ``no'' texts leads to improvements of AUROC by 6.83\%, 6.85\%, 0.66\%, and 6.86\% across the four datasets, compared to handcrafted texts. Regarding CLIPN-A, the incorporation of learnable ``no'' texts results in AUROC improvements of 2.52\%, 2.38\%, and 2.63\% for three datasets, while showing a decrease of 1.22\% AUROC for the Texture dataset, as compared to handcrafted texts.

\begin{table}[t]
    \centering
    \caption{Ablation study on the handcrafted and learnable ``no'' texts. $\star$ and $\dag$ refer to our method uses the handcrafted and learnable  ``no'' prompts, respectively. CLIPN-C and CLIPN-A indicate CLIPN with competing-to-win and agreeing-to-differ algorithms, respectively.}
    \label{tab:prompt}

\begin{tabular}{c|cccc}
\toprule
Model & iNaturalist & SUN & Texture & Places  \\
\midrule
CLIPN-C$\star$    &  84.05   &    82.53   &    77.62     &     79.99 \\
CLIPN-C$\dag$    &  90.88   &    89.38  &    78.28    &     86.85 \\
$\vartriangle$ & 6.83 & 6.85 & 0.66 & 6.86 \\
\midrule
CLIPN-A$\star$    &  92.75   &    91.55   &    92.15     &      89.65 \\
CLIPN-A$\dag$    &  95.27    &    93.93   &    90.93     &      92.28 \\
$\vartriangle$ & 2.52 & 2.38 & -1.22 & 2.63 \\
\bottomrule
\end{tabular}
\end{table}

\begin{figure}[htbp]
    \centering
    \subfigure[Results of MSP $thres=0.5$]{
        \includegraphics[width=0.22\textwidth]{"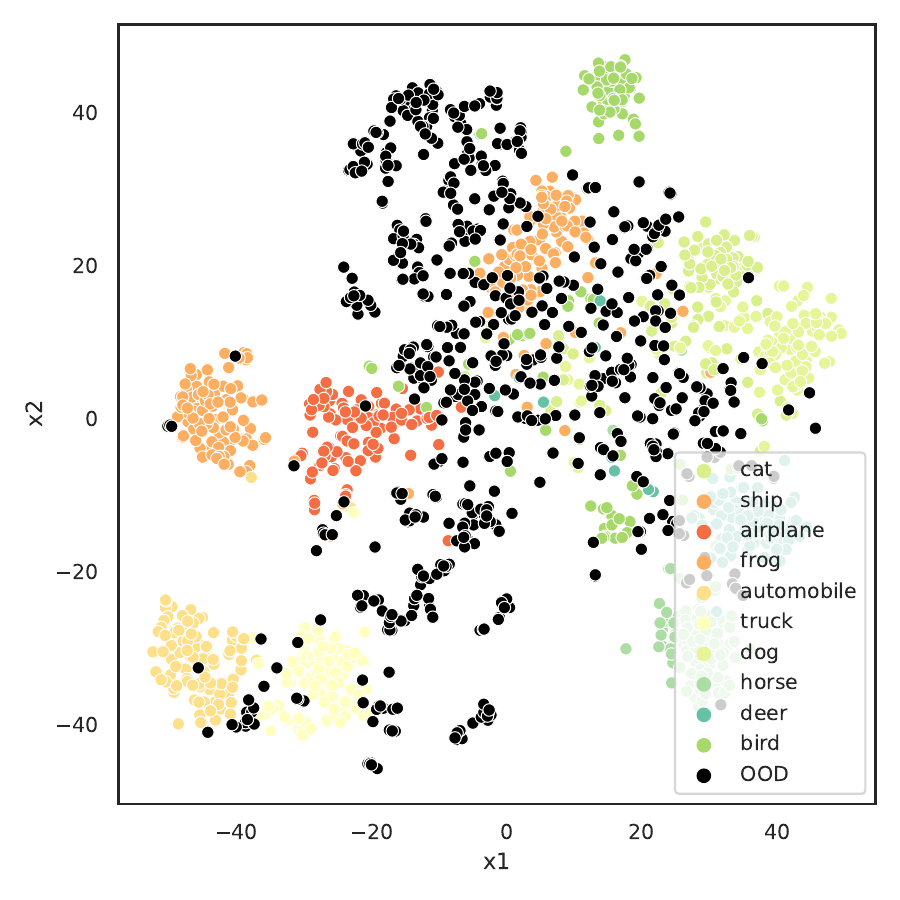"}
    }
    \subfigure[Results of MSP $thres=0.7$]{
        \includegraphics[width=0.22\textwidth]{"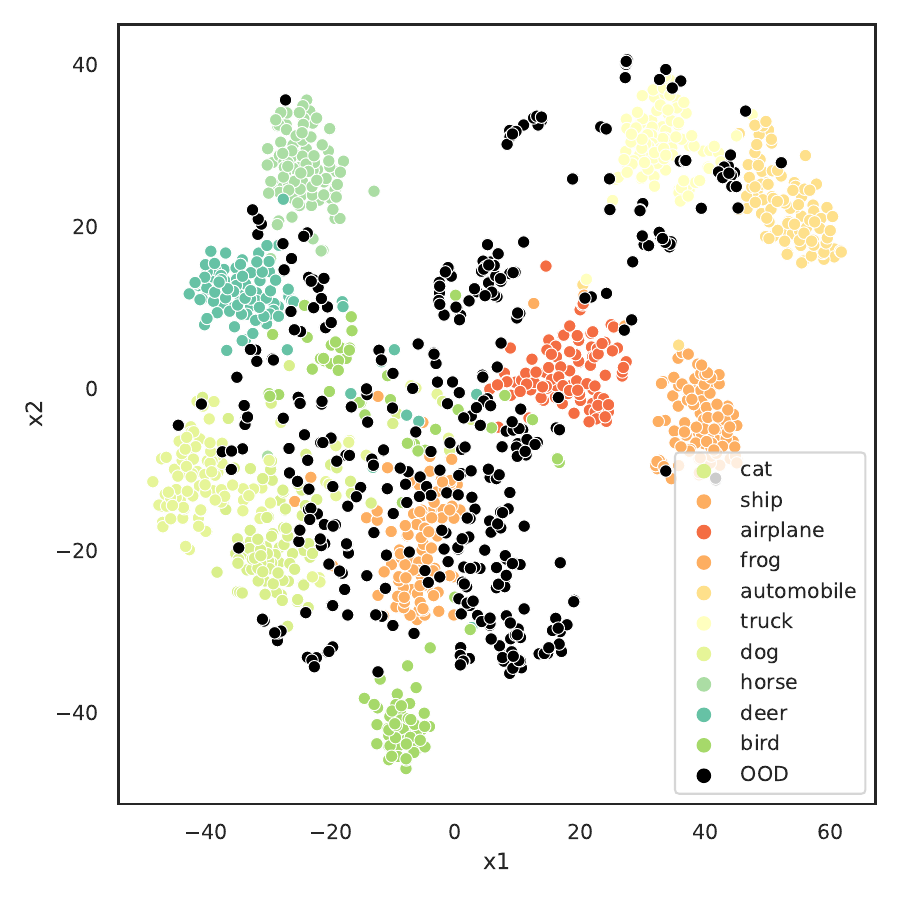"}
    }
    \subfigure[Results of MSP $thres=0.9$]{
        \includegraphics[width=0.22\textwidth]{"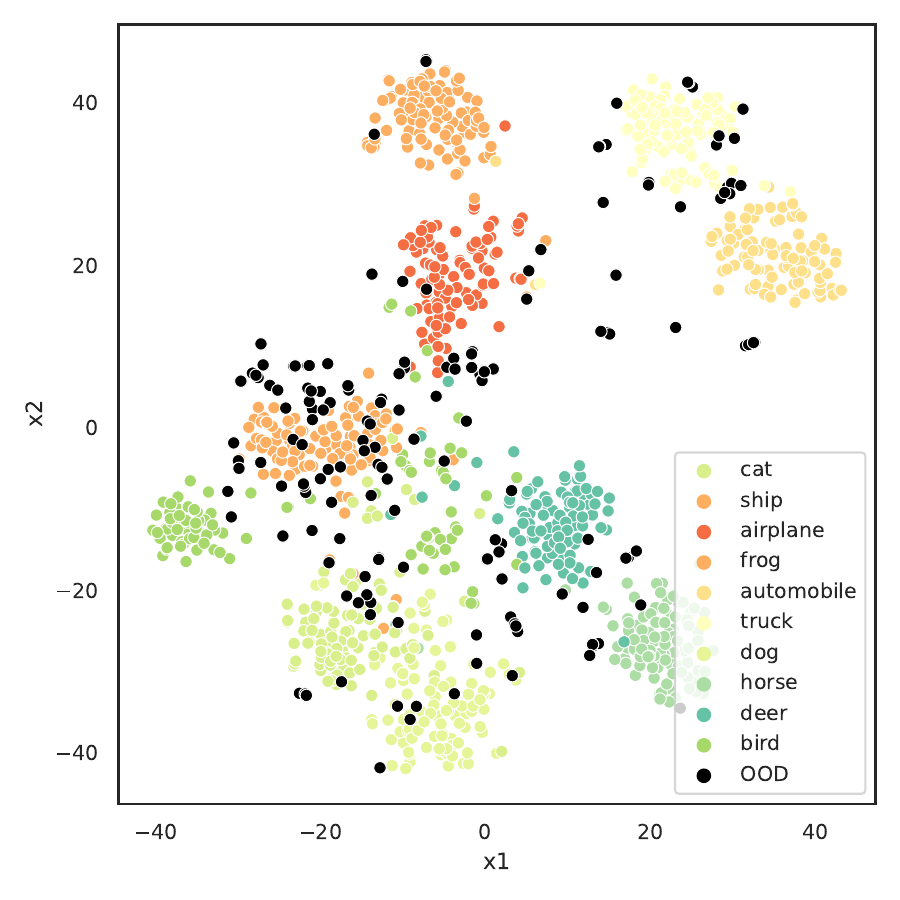"}
    }
    \subfigure[Results of ours]{
        \includegraphics[width=0.22\textwidth]{"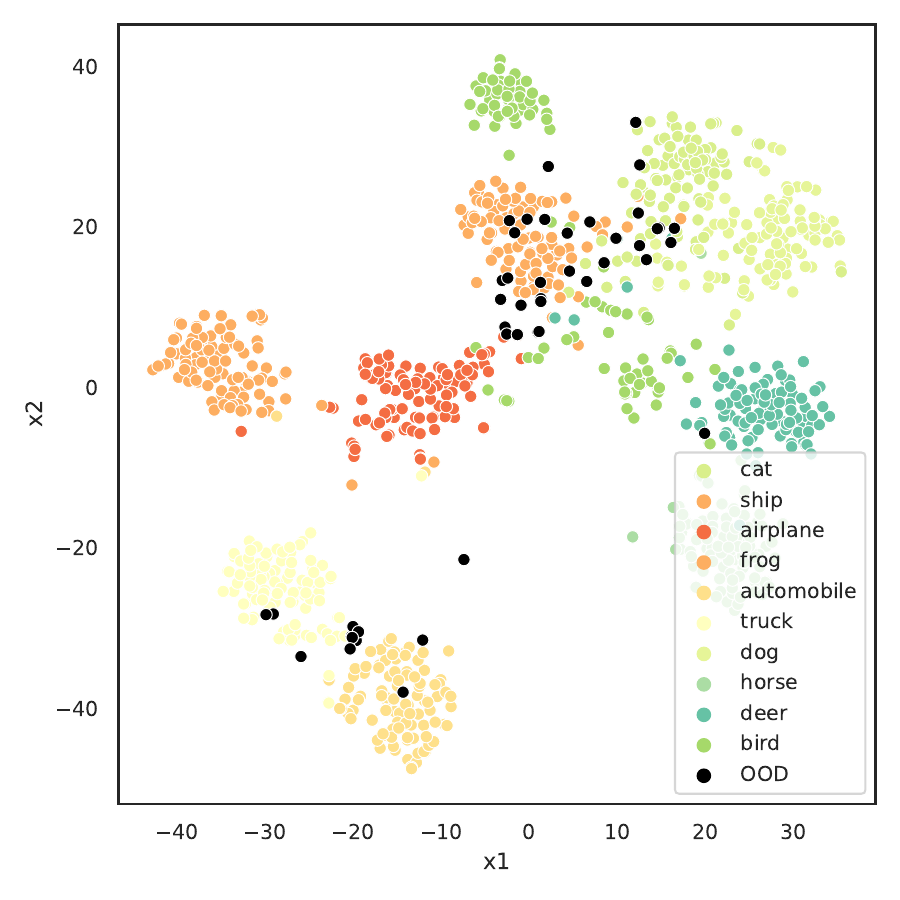"}
    }
    \caption{Mis-classified OOD (dark dots) and ID (other colors) feature visualization via T-SNE. (a-c) Results of MSP~\cite{hendrycks2016baseline} with threshold [0.5, 0.7, 0.9]. (d) Result of our CLIPN-A. The ID and OOD features are captured from images in CIFAR-10 and CIFAR-100, using Eqn.~\ref{eqn:1}. The misclassified samples indicate that samples were classified incorrectly as CIFAR-10 classes, either by MSP or by ours. Besides, the number of mis-classified OOD features are $666$, $410$, $171$, and $45$ for (a), (b), (c), and (d), respectively. }
    \label{fig:tsne}
\end{figure}

\noindent \textbf{The effectiveness to eliminate the mis-classified OOD samples}.
We evaluate MSP and CLIPN-A on CIFAR-10 (ID dataset) and CIFAR-100 (OOD dataset), then use T-SNE~\cite{tsne} to visualize the features (using Eqn.~\ref{eqn:1}) of ID samples and mis-classified OOD samples (belonging to OOD yet classified as ID) caused by MSP and ours. 
The results are shown in Figure.~\ref{fig:tsne}. For MSP, the numbers of mis-classified OOD samples are $666$, $410$, and $171$ under the threshold $0.5$, $0.7$, and $0.9$, which are heavily falling into ID sample clusters. However, our threshold-free CLIPN-A only has $45$ mis-classified OOD samples, significantly less than MSP and rarely fall into ID clusters. It demonstrates that our CLIPN-A can effectively use the ``no'' logic to eliminate mis-classified OOD samples.

\begin{figure}[htbp]
    \centering
    \subfigure[Similarity density of CLIP]{
        \includegraphics[width=0.22\textwidth]{"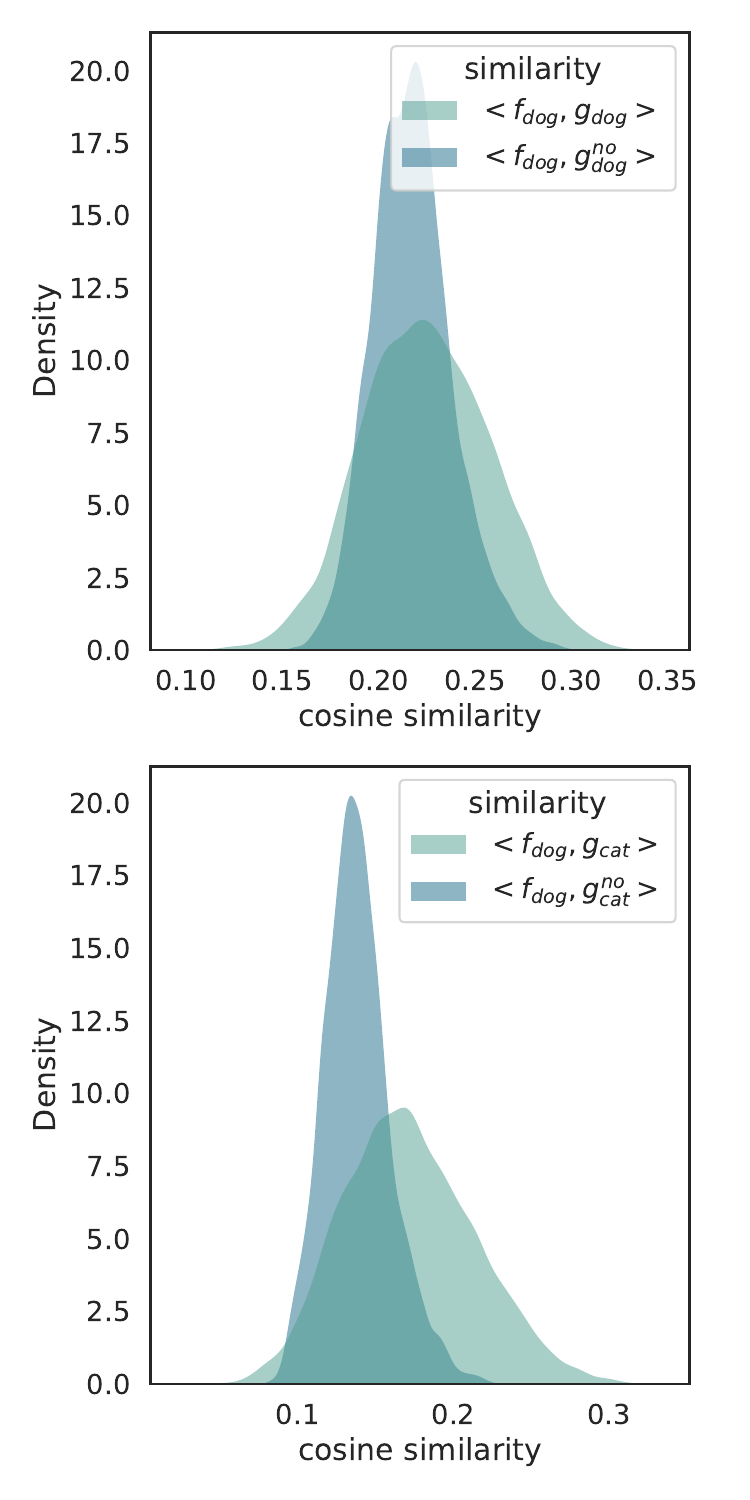"}
    }
    \subfigure[Similarity density of CLIPN]{
	\includegraphics[width=0.22\textwidth]{"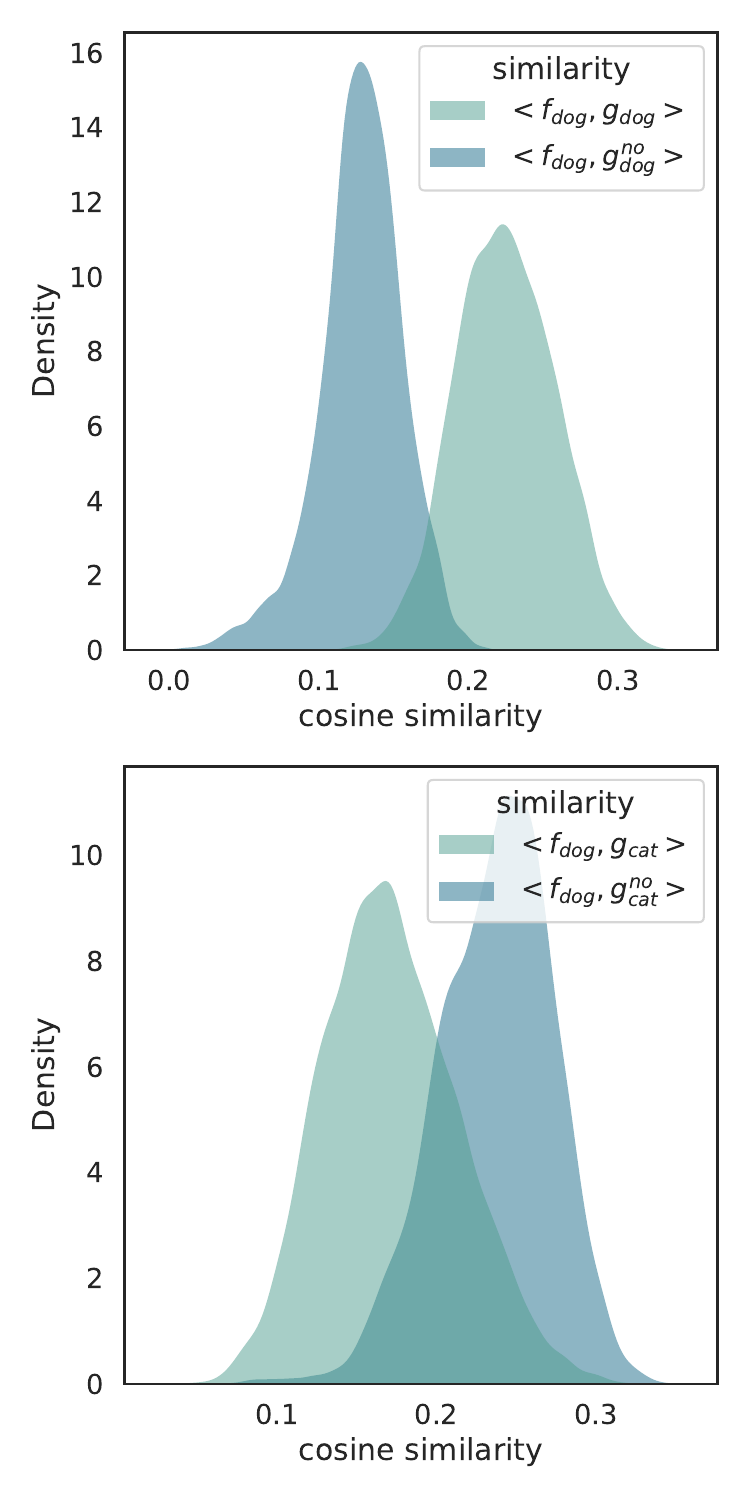"}
    }
    \caption{Similarity density of CLIP and CLIPN. (a) and (b) show the similarity density between dog images in CIFAR-10 and four kind of texts in terms of CLIP and CLIPN, respectively.}
    \label{fig:dog_clipn}
\end{figure}

\noindent \textbf{The capacity to match images and ``no'' texts}. 
We conduct an experiment to evaluate the capacity of matching images and ``no'' texts in terms of  CLIP and CLIPN. Specifically, given CLIP or CLIPN, we extract all dog features from CIFAR-10 (for each dog image, we get the image feature $\bm{f}_{dog}$ using Eqn.~\ref{eqn:1}). Then we use Eqn.~\ref{eqn:1} to get four text features, $\bm{g}_{dog}$, $\bm{g}_{cat}$ (standard dog/cat texts) and $\bm{g}^{no}_{dog}$, $\bm{g}^{no}_{cat}$ (``no'' dog/cat texts).
Next we calculate the two group cosine similarities $<\bm{f}_{dog}, \bm{g}_{dog}>$ and $<\bm{f}_{dog}, \bm{g}^{no}_{dog}>$; $<\bm{f}_{dog}, \bm{g}_{cat}>$ and $<\bm{f}_{dog}, \bm{g}^{no}_{cat}>$. Finally, we use the kernel density estimation function~\cite{scikit-learn} to estimate the density of two group similarities for CLIP and CLIPN, as shown in Figure.~\ref{fig:dog_clipn}. 
Followed the match-ness defined in Eqn.~\ref{eqn:m}, the ideal outcome should be that dog features are matched with dog texts and ``no'' cat texts. From Figure.~\ref{fig:dog_clipn}, we find that the original CLIP fails to achieve it because all similarities are mixed together. Conversely, similarity density of CLIPN has obviously higher value on $<\bm{f}_{dog}, \bm{g}_{dog}>$ and lower $<\bm{f}_{dog}, \bm{g}^{no}_{cat}>$, implying CLIPN has strong ability to match images and ``no'' texts.

\noindent \textbf{Computational and storage cost of the training and inference periods.} Our CLIPN significantly reduces both training time and GPU memory usage when compared to training CLIP, all while achieving substantial performance improvements with minimal additional cost. Specifically, we employ four commonly used metrics to quantify these benefits: (1) Floating-Point Operations per Second (FLOPs); (2) Number of Parameters; (3) Training Time per Iteration; (4) GPU Memory Usage per Iteration. Metric (1) is calculated solely during the forward step, while metrics (3) and (4) pertain to the computation and memory usage during both the forward and backward steps.

\begin{table}[h!]
\caption{Cost comparison between CLIP and CLIPN during the training period. \textcolor{red}{Red} presents the FLOPs and parameters that need backward computations. FLOPs/parameters are decomposed into sub-parts of image, text, and ``no'' text encoders. \textcolor{blue}{Blue} indicates the percentage of the decreased cost.
All experiments are evaluated on a device with 4$\times$RTX3090 (24G) and 2$\times$Intel Xeon Gold 5118.  }
\label{tab:1}
\resizebox{0.48\textwidth}{!}{
\begin{tabular}{ccccc}
\toprule
\multicolumn{1}{c|}{Method} & FLOPs (G) & Parameters (M) & Time/Iter (s) & GPU Usage (G) \\ 
\midrule
\multicolumn{5}{c}{ViT-B-32 (Input size: \textbf{16}x3x224x224) }                           \\ 
\midrule
\multicolumn{1}{c|}{CLIP} &  \textcolor{red}{141 + 95} &  \textcolor{red}{87.9 + 37.8}  & 0.22 & 5.86 \\
\multicolumn{1}{c|}{CLIPN (ours)} &  141 + 95 \textcolor{red}{ + 95} &  87.9 + 37.8 \textcolor{red}{ + 37.8} & 0.14  ($\downarrow$ \textcolor{blue}{36.4\%}) & 4.16 ($\downarrow$ \textcolor{blue}{29.0\%}) \\
\midrule
\multicolumn{5}{c}{ViT-L-14 (Input size: \textbf{16}x3x224x224)}     \\
\midrule
\multicolumn{1}{c|}{CLIP}   &  \textcolor{red}{2594 + 213} &  \textcolor{red}{304.0 + 85.1}  & 0.58 & 18.91 \\
\multicolumn{1}{c|}{CLIPN (ours)}   &  2594 + 213 \textcolor{red}{ + 213} &  304.0 + 85.1 \textcolor{red}{+ 85.1}  & 0.24 ($\downarrow$ \textcolor{blue}{58.6\%}) & 6.62 ($\downarrow$ \textcolor{blue}{65.0\%}) \\
\midrule
\multicolumn{5}{c}{ViT-L-14 (Input size: \textbf{64}x3x224x224)}     \\
\midrule
\multicolumn{1}{c|}{CLIP}   &  \textcolor{red}{10375 + 852} &  \textcolor{red}{304.0 + 85.1}  & - & OOM ($\gg$ 24) \\
\multicolumn{1}{c|}{CLIPN (ours)}   &  10375 + 852 \textcolor{red}{+ 852} &  304.0 + 85.1\textcolor{red}{+ 85.1}  & 0.49 & 8.31 \\
\bottomrule
\end{tabular}
}
\end{table}

\begin{table}[h!]

\caption{Cost and performance comparison between CLIP and CLIPN during the inference period. The performance of AUROC is CLIP with MaxLogits and CLIPN-A. The costs are calculated under the same set of ViT-B-32 as Table.~\ref{tab:1}.  }
\label{tab:2}
\resizebox{0.48\textwidth}{!}{
\begin{tabular}{ccccc}
\toprule
\multicolumn{1}{c|}{Method} & FLOPs (G) & Parameters (M) & Avg AUROC & Improved AUROC\\                  
\midrule
\multicolumn{1}{c|}{CLIP} & 141 + 0.016 &  87.9 + 0.51 & 85.59\% & - \\
\multicolumn{1}{c|}{CLIPN (ours)} &  141 + 0.016 + 0.016 & 87.9 + 0.51 + 0.51 & 90.53\% & $\uparrow$ \textcolor{red}{4.94\%} \\
\bottomrule
\end{tabular}
}
\end{table}

In terms of the training period, as indicated in Table~\ref{tab:1}, CLIPN demonstrates reductions of \textcolor{blue}{$36.4\%$} and \textcolor{blue}{$29.0\%$} in training time per iteration and GPU usage on ViT-B-32, respectively. Additionally, on ViT-L-14, CLIPN achieves reductions of \textcolor{blue}{$58.6\%$} and \textcolor{blue}{$65.0\%$} in training time per iteration and GPU usage when compared to CLIP. This outcome is attributed to the decreased number of Floating-Point Operations (FLOPs) and parameters during backward computation. Furthermore, CLIPN enables the training of ViT-L-14 on the RTX3090 with a comparatively large batch size.
Regarding the inference period, outlined in Table~\ref{tab:2}, the conversion of two text encoders to classifiers results in a negligible increase in computation and parameters, specifically by $0.016$G and $0.51$M, respectively. However, this minor adjustment leads to an average AUROC increase of 4.94\%.

\section{Conclusion and Limitation}
\label{sec:conclude}

This paper presents a novel framework, namely CLIPN, for OOD detection by teaching CLIP to say ``no''. The key insight is to equip CLIP with the capability of distinguishing OOD and ID samples via positive-semantic prompts and negation-semantic prompts. Specifically, we propose new ``no'' prompts and text encoder. Further, we propose two training losses: an image-text binary-opposite loss and a text semantic-opposite loss. These losses enable CLIP to recognize scenarios where it should respond with ``no" and to understand the meaning of "no''.
Additionally, we propose two threshold-free inference algorithms: competing-to-win and agreeing-to-differ. 
Extensive experimental results demonstrated the effectiveness of our method. 
One limitation of our approach is the lack of clear demonstrations of its extension to OOD segmentation or detection tasks. Another limitation of our approach is the uncertainty regarding its effectiveness for OOD classification in specialized datasets such as medical and satellite images. This is primarily because our model is based on CLIP, and its effectiveness in specialized datasets is still underexplored. 

\section{Acknowledgement}
This work is supported by grants from Foshan HKUST Projects (Grants FSUST21-HKUST10E and FSUST21-HKUST11E), the Hong Kong Innovation and Technology Fund (Projects ITS/030/21), and the Beijing Institute of Collaborative Innovation (BICI) in collaboration with HKUST (Grant HCIC-004).

\clearpage

{\small
\bibliographystyle{ieee_fullname}
\bibliography{egbib}

\begin{thebibliography}{10}\itemsep=-1pt

\bibitem{bai2023effectiveness}
Jianhong Bai, Zuozhu Liu, Hualiang Wang, Jin Hao, Yang Feng, Huanpeng Chu, and
  Haoji Hu.
\newblock On the effectiveness of out-of-distribution data in self-supervised
  long-tail learning.
\newblock {\em arXiv preprint arXiv:2306.04934}, 2023.

\bibitem{bendale2016towards}
Abhijit Bendale and Terrance~E Boult.
\newblock Towards open set deep networks.
\newblock In {\em Proceedings of the IEEE conference on computer vision and
  pattern recognition}, pages 1563--1572, 2016.

\bibitem{chen2017outlier}
Jinghui Chen, Saket Sathe, Charu Aggarwal, and Deepak Turaga.
\newblock Outlier detection with autoencoder ensembles.
\newblock In {\em Proceedings of the 2017 SIAM international conference on data
  mining}, pages 90--98. SIAM, 2017.

\bibitem{devlin2018bert}
Jacob Devlin, Ming-Wei Chang, Kenton Lee, and Kristina Toutanova.
\newblock Bert: Pre-training of deep bidirectional transformers for language
  understanding.
\newblock {\em arXiv preprint arXiv:1810.04805}, 2018.

\bibitem{ding2021support}
Xinpeng Ding, Nannan Wang, Shiwei Zhang, De Cheng, Xiaomeng Li, Ziyuan Huang,
  Mingqian Tang, and Xinbo Gao.
\newblock Support-set based cross-supervision for video grounding.
\newblock In {\em Proceedings of the IEEE/CVF International Conference on
  Computer Vision}, pages 11573--11582, 2021.

\bibitem{dosovitskiy2020image}
Alexey Dosovitskiy, Lucas Beyer, Alexander Kolesnikov, Dirk Weissenborn,
  Xiaohua Zhai, Thomas Unterthiner, Mostafa Dehghani, Matthias Minderer, Georg
  Heigold, Sylvain Gelly, et~al.
\newblock An image is worth 16x16 words: Transformers for image recognition at
  scale.
\newblock {\em arXiv preprint arXiv:2010.11929}, 2020.

\bibitem{drummond2006open}
Nick Drummond and Rob Shearer.
\newblock The open world assumption.
\newblock In {\em eSI Workshop: The Closed World of Databases meets the Open
  World of the Semantic Web}, volume~15, page~1, 2006.

\bibitem{esmaeilpour2022zero}
Sepideh Esmaeilpour, Bing Liu, Eric Robertson, and Lei Shu.
\newblock Zero-shot out-of-distribution detection based on the pretrained model
  clip.
\newblock In {\em Proceedings of the AAAI conference on artificial
  intelligence}, 2022.

\bibitem{fort2021exploring}
Stanislav Fort, Jie Ren, and Balaji Lakshminarayanan.
\newblock Exploring the limits of out-of-distribution detection.
\newblock {\em Advances in Neural Information Processing Systems},
  34:7068--7081, 2021.

\bibitem{gao2021clip}
Peng Gao, Shijie Geng, Renrui Zhang, Teli Ma, Rongyao Fang, Yongfeng Zhang,
  Hongsheng Li, and Yu Qiao.
\newblock Clip-adapter: Better vision-language models with feature adapters.
\newblock {\em arXiv preprint arXiv:2110.04544}, 2021.

\bibitem{gao2019dynamic}
Peng Gao, Zhengkai Jiang, Haoxuan You, Pan Lu, Steven~CH Hoi, Xiaogang Wang,
  and Hongsheng Li.
\newblock Dynamic fusion with intra-and inter-modality attention flow for
  visual question answering.
\newblock In {\em Proceedings of the IEEE/CVF conference on computer vision and
  pattern recognition}, pages 6639--6648, 2019.

\bibitem{he2016deep}
Kaiming He, Xiangyu Zhang, Shaoqing Ren, and Jian Sun.
\newblock Deep residual learning for image recognition.
\newblock In {\em Proceedings of the IEEE conference on computer vision and
  pattern recognition}, pages 770--778, 2016.

\bibitem{hendrycks2019scaling}
Dan Hendrycks, Steven Basart, Mantas Mazeika, Mohammadreza Mostajabi, Jacob
  Steinhardt, and Dawn Song.
\newblock Scaling out-of-distribution detection for real-world settings.
\newblock {\em arXiv preprint arXiv:1911.11132}, 2019.

\bibitem{hendrycks2016baseline}
Dan Hendrycks and Kevin Gimpel.
\newblock A baseline for detecting misclassified and out-of-distribution
  examples in neural networks.
\newblock {\em arXiv preprint arXiv:1610.02136}, 2016.

\bibitem{huynh2020fine}
Dat Huynh and Ehsan Elhamifar.
\newblock Fine-grained generalized zero-shot learning via dense attribute-based
  attention.
\newblock In {\em Proceedings of the IEEE/CVF conference on computer vision and
  pattern recognition}, pages 4483--4493, 2020.

\bibitem{ilharco_gabriel_2021_5143773}
Gabriel Ilharco, Mitchell Wortsman, Ross Wightman, Cade Gordon, Nicholas
  Carlini, Rohan Taori, Achal Dave, Vaishaal Shankar, Hongseok Namkoong, John
  Miller, Hannaneh Hajishirzi, Ali Farhadi, and Ludwig Schmidt.
\newblock Openclip, July 2021.
\newblock If you use this software, please cite it as below.

\bibitem{kim2018bilinear}
Jin-Hwa Kim, Jaehyun Jun, and Byoung-Tak Zhang.
\newblock Bilinear attention networks.
\newblock {\em Advances in neural information processing systems}, 31, 2018.

\bibitem{koner2021oodformer}
Rajat Koner, Poulami Sinhamahapatra, Karsten Roscher, Stephan G{\"u}nnemann,
  and Volker Tresp.
\newblock Oodformer: Out-of-distribution detection transformer.
\newblock {\em arXiv preprint arXiv:2107.08976}, 2021.

\bibitem{krizhevsky2009learning}
Alex Krizhevsky, Geoffrey Hinton, et~al.
\newblock Learning multiple layers of features from tiny images.
\newblock 2009.

\bibitem{krizhevsky2017imagenet}
Alex Krizhevsky, Ilya Sutskever, and Geoffrey~E Hinton.
\newblock Imagenet classification with deep convolutional neural networks.
\newblock {\em Communications of the ACM}, 60(6):84--90, 2017.

\bibitem{lee2018simple}
Kimin Lee, Kibok Lee, Honglak Lee, and Jinwoo Shin.
\newblock A simple unified framework for detecting out-of-distribution samples
  and adversarial attacks.
\newblock {\em Advances in neural information processing systems}, 31, 2018.

\bibitem{li2023clip}
Yi Li, Hualiang Wang, Yiqun Duan, and Xiaomeng Li.
\newblock Clip surgery for better explainability with enhancement in
  open-vocabulary tasks.
\newblock {\em arXiv preprint arXiv:2304.05653}, 2023.

\bibitem{li2022exploring}
Yi Li, Hualiang Wang, Yiqun Duan, Hang Xu, and Xiaomeng Li.
\newblock Exploring visual interpretability for contrastive language-image
  pre-training.
\newblock {\em arXiv preprint arXiv:2209.07046}, 2022.

\bibitem{li2022freeseg}
Yi Li, Huifeng Yao, Hualiang Wang, and Xiaomeng Li.
\newblock Freeseg: Free mask from interpretable contrastive language-image
  pretraining for semantic segmentation.
\newblock {\em arXiv preprint arXiv:2209.13558}, 2022.

\bibitem{liang2017enhancing}
Shiyu Liang, Yixuan Li, and Rayadurgam Srikant.
\newblock Enhancing the reliability of out-of-distribution image detection in
  neural networks.
\newblock {\em arXiv preprint arXiv:1706.02690}, 2017.

\bibitem{liu2023chatgpt}
Jiaxiang Liu, Tianxiang Hu, Yan Zhang, Xiaotang Gai, Yang Feng, and Zuozhu Liu.
\newblock A chatgpt aided explainable framework for zero-shot medical image
  diagnosis.
\newblock {\em arXiv preprint arXiv:2307.01981}, 2023.

\bibitem{liu2020energy}
Weitang Liu, Xiaoyun Wang, John Owens, and Yixuan Li.
\newblock Energy-based out-of-distribution detection.
\newblock {\em Advances in Neural Information Processing Systems},
  33:21464--21475, 2020.

\bibitem{liu2021swin}
Ze Liu, Yutong Lin, Yue Cao, Han Hu, Yixuan Wei, Zheng Zhang, Stephen Lin, and
  Baining Guo.
\newblock Swin transformer: Hierarchical vision transformer using shifted
  windows.
\newblock In {\em Proceedings of the IEEE/CVF International Conference on
  Computer Vision}, pages 10012--10022, 2021.

\bibitem{lu2019vilbert}
Jiasen Lu, Dhruv Batra, Devi Parikh, and Stefan Lee.
\newblock Vilbert: Pretraining task-agnostic visiolinguistic representations
  for vision-and-language tasks.
\newblock {\em Advances in neural information processing systems}, 32, 2019.

\bibitem{mcm}
Yifei Ming, Ziyang Cai, Jiuxiang Gu, Yiyou Sun, Wei Li, and Yixuan Li.
\newblock Delving into out-of-distribution detection with vision-language
  representations.
\newblock {\em arXiv preprint arXiv:2211.13445}, 2022.

\bibitem{ndiour2020out}
Ibrahima Ndiour, Nilesh Ahuja, and Omesh Tickoo.
\newblock Out-of-distribution detection with subspace techniques and
  probabilistic modeling of features.
\newblock {\em arXiv preprint arXiv:2012.04250}, 2020.

\bibitem{nguyen2015deep}
Anh Nguyen, Jason Yosinski, and Jeff Clune.
\newblock Deep neural networks are easily fooled: High confidence predictions
  for unrecognizable images.
\newblock In {\em Proceedings of the IEEE conference on computer vision and
  pattern recognition}, pages 427--436, 2015.

\bibitem{scikit-learn}
F. Pedregosa, G. Varoquaux, A. Gramfort, V. Michel, B. Thirion, O. Grisel, M.
  Blondel, P. Prettenhofer, R. Weiss, V. Dubourg, J. Vanderplas, A. Passos, D.
  Cournapeau, M. Brucher, M. Perrot, and E. Duchesnay.
\newblock Scikit-learn: Machine learning in {P}ython.
\newblock {\em Journal of Machine Learning Research}, 12:2825--2830, 2011.

\bibitem{petroni2019language}
Fabio Petroni, Tim Rockt{\"a}schel, Patrick Lewis, Anton Bakhtin, Yuxiang Wu,
  Alexander~H Miller, and Sebastian Riedel.
\newblock Language models as knowledge bases?
\newblock {\em arXiv preprint arXiv:1909.01066}, 2019.

\bibitem{radford2021learning}
Alec Radford, Jong~Wook Kim, Chris Hallacy, Aditya Ramesh, Gabriel Goh,
  Sandhini Agarwal, Girish Sastry, Amanda Askell, Pamela Mishkin, Jack Clark,
  et~al.
\newblock Learning transferable visual models from natural language
  supervision.
\newblock In {\em International Conference on Machine Learning}, pages
  8748--8763. PMLR, 2021.

\bibitem{schuhmann2022laion}
Christoph Schuhmann, Romain Beaumont, Richard Vencu, Cade Gordon, Ross
  Wightman, Mehdi Cherti, Theo Coombes, Aarush Katta, Clayton Mullis, Mitchell
  Wortsman, et~al.
\newblock Laion-5b: An open large-scale dataset for training next generation
  image-text models.
\newblock {\em arXiv preprint arXiv:2210.08402}, 2022.

\bibitem{sharma2018conceptual}
Piyush Sharma, Nan Ding, Sebastian Goodman, and Radu Soricut.
\newblock Conceptual captions: A cleaned, hypernymed, image alt-text dataset
  for automatic image captioning.
\newblock In {\em Proceedings of ACL}, 2018.

\bibitem{sun2021react}
Yiyou Sun, Chuan Guo, and Yixuan Li.
\newblock React: Out-of-distribution detection with rectified activations.
\newblock {\em Advances in Neural Information Processing Systems}, 34:144--157,
  2021.

\bibitem{tan2019lxmert}
Hao Tan and Mohit Bansal.
\newblock Lxmert: Learning cross-modality encoder representations from
  transformers.
\newblock {\em arXiv preprint arXiv:1908.07490}, 2019.

\bibitem{tsne}
Laurens Van~der Maaten and Geoffrey Hinton.
\newblock Visualizing data using t-sne.
\newblock {\em Journal of machine learning research}, 9(11), 2008.

\bibitem{van2018inaturalist}
Grant Van~Horn, Oisin Mac~Aodha, Yang Song, Yin Cui, Chen Sun, Alex Shepard,
  Hartwig Adam, Pietro Perona, and Serge Belongie.
\newblock The inaturalist species classification and detection dataset.
\newblock In {\em Proceedings of the IEEE conference on computer vision and
  pattern recognition}, pages 8769--8778, 2018.

\bibitem{wang2022vim}
Haoqi Wang, Zhizhong Li, Litong Feng, and Wayne Zhang.
\newblock Vim: Out-of-distribution with virtual-logit matching.
\newblock In {\em Proceedings of the IEEE/CVF Conference on Computer Vision and
  Pattern Recognition}, pages 4921--4930, 2022.

\bibitem{wang2018zero}
Xiaolong Wang, Yufei Ye, and Abhinav Gupta.
\newblock Zero-shot recognition via semantic embeddings and knowledge graphs.
\newblock In {\em Proceedings of the IEEE conference on computer vision and
  pattern recognition}, pages 6857--6866, 2018.

\bibitem{wortsman2022robust}
Mitchell Wortsman, Gabriel Ilharco, Jong~Wook Kim, Mike Li, Simon Kornblith,
  Rebecca Roelofs, Raphael~Gontijo Lopes, Hannaneh Hajishirzi, Ali Farhadi,
  Hongseok Namkoong, et~al.
\newblock Robust fine-tuning of zero-shot models.
\newblock In {\em Proceedings of the IEEE/CVF Conference on Computer Vision and
  Pattern Recognition}, pages 7959--7971, 2022.

\bibitem{xiao2010sun}
Jianxiong Xiao, James Hays, Krista~A Ehinger, Aude Oliva, and Antonio Torralba.
\newblock Sun database: Large-scale scene recognition from abbey to zoo.
\newblock In {\em 2010 IEEE computer society conference on computer vision and
  pattern recognition}, pages 3485--3492. IEEE, 2010.

\bibitem{xu2022groupvit}
Jiarui Xu, Shalini De~Mello, Sifei Liu, Wonmin Byeon, Thomas Breuel, Jan Kautz,
  and Xiaolong Wang.
\newblock Groupvit: Semantic segmentation emerges from text supervision.
\newblock In {\em Proceedings of the IEEE/CVF Conference on Computer Vision and
  Pattern Recognition}, pages 18134--18144, 2022.

\bibitem{yu2019deep}
Zhou Yu, Jun Yu, Yuhao Cui, Dacheng Tao, and Qi Tian.
\newblock Deep modular co-attention networks for visual question answering.
\newblock In {\em Proceedings of the IEEE/CVF conference on computer vision and
  pattern recognition}, pages 6281--6290, 2019.

\bibitem{zhou2017places}
Bolei Zhou, Agata Lapedriza, Aditya Khosla, Aude Oliva, and Antonio Torralba.
\newblock Places: A 10 million image database for scene recognition.
\newblock {\em IEEE transactions on pattern analysis and machine intelligence},
  40(6):1452--1464, 2017.

\bibitem{zhou2022conditional}
Kaiyang Zhou, Jingkang Yang, Chen~Change Loy, and Ziwei Liu.
\newblock Conditional prompt learning for vision-language models.
\newblock In {\em Proceedings of the IEEE/CVF Conference on Computer Vision and
  Pattern Recognition}, pages 16816--16825, 2022.

\bibitem{zhou2022learning}
Kaiyang Zhou, Jingkang Yang, Chen~Change Loy, and Ziwei Liu.
\newblock Learning to prompt for vision-language models.
\newblock {\em International Journal of Computer Vision}, 130(9):2337--2348,
  2022.

\end{thebibliography}
}

\end{document}